\documentclass[runningheads]{llncs}
\usepackage{graphicx}
\usepackage{amsmath}
\usepackage{amsmath,amssymb,amsfonts}
\usepackage{graphicx}
\usepackage{textcomp}
\usepackage{xcolor}

\usepackage{algorithm}
\usepackage{algorithmic}

\usepackage{blindtext}
\usepackage{enumitem}

\usepackage{booktabs} 

\usepackage[T1]{fontenc} 

\DeclareMathOperator{\ensa}{\textsc{ens1}}
\DeclareMathOperator{\ensb}{\textsc{ens2}}
\DeclareMathOperator{\ensc}{\textsc{ens3}}
\DeclareMathOperator{\ensd}{\textsc{ens4}}
\DeclareMathOperator{\ense}{\textsc{ens5}}
\DeclareMathOperator{\ensf}{\textsc{ens6}}
\DeclareMathOperator{\ensg}{\textsc{ens7}}
\DeclareMathOperator{\randnn}{\textsc{randNN}}
\DeclareMathOperator{\randnns}{\textsc{randNN}s}

\begin{document}
\title{Ensembles of Randomized Neural Networks for Pattern-based Time Series Forecasting\thanks{Supported by Grant 2017/27/B/ST6/01804 from the National Science Centre, Poland.}}
\titlerunning{Ensembles of Randomized NNs for Pattern-based Time Series Forecasting}
%
\author{Grzegorz Dudek\orcidID{0000-0002-2285-0327} \and
Paweł Pełka\orcidID{0000-0002-2609-811X}}
\authorrunning{G. Dudek, P. Pełka}
%
\institute{Electrical Engineering Faculty, Częstochowa University of Technology,\\ Częstochowa, Poland\\
\email{\{grzegorz.dudek,pawel.pelka\}@pcz.pl}}

\maketitle              
\begin{abstract}

In this work, we propose an ensemble forecasting approach based on randomized neural networks. Improved randomized learning streamlines the fitting abilities of individual learners by generating network parameters in accordance with the data and target function features. A pattern-based representation of time series makes the proposed approach suitable for forecasting time series with multiple seasonality. We propose six strategies for controlling the diversity of ensemble members. Case studies conducted on four real-world forecasting problems verified the effectiveness and superior performance of the proposed ensemble forecasting approach. It outperformed statistical models as well as state-of-the-art machine learning models in terms of forecasting accuracy. The proposed approach has several advantages: fast and easy training, simple architecture, ease of implementation, high accuracy and the ability to deal with nonstationarity and multiple seasonality in time series.      

\end{abstract}

\section{Introduction}

Time series (TS) forecasting plays an important role in  many fields, including commerce, industry, public administration, politics, health, medicine, etc. \cite{Pal05}. TS expressing different phenomena and processes may include the nonlinear trend, multiple seasonal cycles of different lengths and random fluctuations. This makes the relationship between predictors and output variables very complex and places high demands on forecasting models. Over the years, many sophisticated forecasting models have been proposed including statistical, machine learning (ML) and hybrid solutions.

Among ML models, neural networks (NNs) are the most commonly used. There are a huge number of forecasting models based on different NN architectures \cite{Benidis20}. 
In addition to classic NNs such as multilayer perceptron (MLP), radial basis function NN, generalized regression NN (GRNN), and self-organizing maps \cite{Dudek16}, many models based on deep learning (DL) have been developed recently.
These are composed of combinations of basic structures, such as MLPs, convolutional NNs, and recurrent NNs (RNNs) \cite{Tor21}, \cite{Hew21}. Their success can be largely attributed to increased model complexity and the ability to perform representation learning and cross-learning.

An effective way to increase the performance of a predictive model is ensembling. Ensemble methods combine in some fashion multiple learning algorithms to produce a common response, hopefully improving accuracy and stability compared to a single learner. 
The challenge in ensemble learning, which to a large extent determines its success, is achieving a good tradeoff between the performance and the diversity of the ensemble members \cite{Ree18}. The more accurate and the more diverse the individual learners are, the better the ensemble performance. Depending on the base model type, the diversity of learners can be achieved using different strategies. For example, the winning submission to the M4 Makridakis forecasting competition, was a hybrid model combining exponential smoothing (ETS) and long-short term memory \cite{Smyl20}, which uses three sources of diversity.  
The first is a stochastic training process, the second is similar to bagging, and the third is training ensemble members using different initial parameters.
In another state-of-the-art DL forecasting model, N-Beats \cite{Oresh20}, in order to achieve diversity, each of the ensemble members is trained using a different random initialization and a different random sequence of batches.

Randomization-based NNs are especially suitable for ensembling as they are highly unstable and extremely fast trained \cite{Ren16}. In randomized NNs, the hidden node parameters are selected randomly and the output weights are tuned using a simple and fast least-squares method. This avoids the difficulties associated with gradient-based learning such as slow convergence, local minima problems, and model uncertainties caused by the initial parameters. Ensemble methods based on randomized NNs were presented recently in several papers. In \cite{Alh14}, an ensemble based on decorrelated random vector
functional link (RVFL) networks was proposed using negative correlation learning to control the trade-off among the bias, variance and covariance in the ensemble learning.
A selective ensemble of randomization-based NNs using successive projections algorithm (SPA) was put forward in \cite{Mes18}. SPA improves diversity by selecting uncorrelated members.
In \cite{Hua21}, to enhance the generalization capacities of the ensemble, a novel framework for building an ensemble model by selecting appropriate representatives from a number of randomization-base models (RVFL or stochastic configuration networks) was proposed. 

In the forecasting domain, ensembles of randomization-based NNs are widely used. Examples include: \cite{Li16}, where ensemble members, which are extreme learning machines, learn on TS decomposed by wavelet transform; \cite{Qui18}, where a hybrid incremental learning approach is presented for ensemble forecasting, which is composed of discrete wavelet transform, empirical mode decomposition and RVFL networks as base learners; \cite{Alm20}, where a new bagging ensemble approach based on NN with random weights for online data stream regression is proposed; \cite{Hu21}, where a data-driven evolutionary ensemble forecasting model is proposed using two optimization objectives (error and diversity) and RVFL as a base learner.

Motivated by the good performance of the randomized NN based ensemble forecasting models mentioned above, and new improvements in randomized learning \cite{Dud20} as well as pattern-based TS representation suitable for TS with multiple seasonality \cite{Dudek16}, this work contributes to the development of forecasting models in the following ways:

\begin{enumerate}
	\item A new ensemble forecasting model based on randomized NNs is presented. Enhanced randomized learning improves the fitting abilities of individual learners. To deal with multiple seasonality and nonstationarity, the model applies pattern representation of TS.
	\item Six strategies for generating ensemble diversity are proposed. Each strategy governs diversity in a different way.   
	\item An experimental study using four real-world datasets demonstrates the superior performance of the proposed ensemble forecasting approach when compared to statistical and machine learning forecasting models.
\end{enumerate}

The remaining sections of this work are organized as follows. Section 2 presents the base forecasting model, $\randnn$. Section 3 describes strategies for generating ensemble diversity. The performance of the proposed ensemble forecasting model is evaluated in Section 4. Concluding remarks are given in Section 5.






\section{Forecasting Model}

\begin{figure}
\centering
\includegraphics[width=0.9\textwidth]{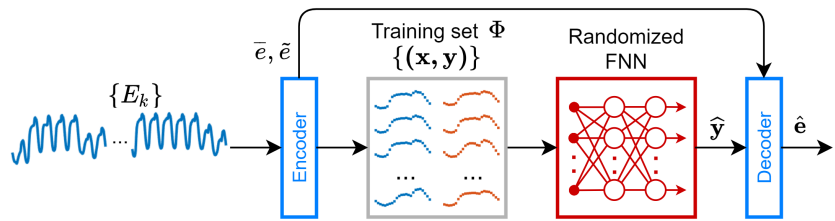}
\caption{Block diagram of the proposed forecasting model.} \label{figModel}
\end{figure}

Fig. \ref{figModel} shows $\randnn$, a base forecasting model, which was designed for forecasting TS with multiple seasonality  \cite{Dud21}. It is used as an ensemble member. $\randnn$ is composed of an encoder, a decoder and a randomized feedforward NN (FNN). 

The encoder transforms the original TS into unified input and output patterns of its seasonal cycles. To create input patterns, the TS expressing multiple seasonality, $\{E_k\}_{k=1}^K$, is divided into seasonal sequences of the shortest length. Let these sequences be expressed by vectors $ \mathbf{e}_i = [E_{i,1}, E_{i,2}, …, E_{i,n}]^T $, where $n$ is the seasonal sequence length and $i=1, 2, ..., K/n$ is the sequence number. These sequences are encoded in input patterns $\mathbf{x}_i = [x_{i,1}, x_{i,2}, …, x_{i,n}]^T$ as follows:

\begin{equation}\label{eq1}
\mathbf{x}_i = \frac{\mathbf{e}_i-\overline{e}_i}{\widetilde{e}_i}
\end{equation}
where $\overline{{e}}_i$ is the mean value of sequence $\mathbf{e}_i$, and $\widetilde{e}_i = \sqrt{\sum_{t=1}^{n} (E_{i,t}-\overline{e}_i)^2}$ is a measure of sequence $\mathbf{e}_i$ dispersion.

Note that the x-patterns are normalized versions of centered vectors $\mathbf{e}_i$. All x-patterns, representing successive seasonal sequences, have a zero mean, the same variance and the same unity length. However, they differ in shape. Thus, the original seasonal sequences, which have a different mean value and dispersion, are unified (see Fig. 2 in \cite{Dud21}). 


The output patterns $\mathbf{y}_i = [y_{i,1}, y_{i,2}, …, y_{i,n}]^T$ represent the forecasted sequences $ \mathbf{e}_{i+\tau} = [E_{i+\tau,1}, E_{i+\tau,2}, …, E_{i+\tau,n}]^T $, where $\tau \geq 1$ is a forecast horizon. The y-patterns are determined as follows:

\begin{equation}\label{eq2}
\mathbf{y}_i = \frac{\mathbf{e}_{i+\tau}-\overline{e}_i}{\widetilde{e}_i}
\end{equation}
where $\overline{{e}}_i$ and $\widetilde{e}_i$ are the same as in \eqref{eq1}.

Note that in \eqref{eq2}, for the $i$-th output pattern, we use the same coding variables $\overline{{e}}_i$ and $\widetilde{e}_i$ as for the $i$-th input pattern. This is because the coding variables for the forecasted sequence, $\overline{e}_{i+\tau}$ and $\widetilde{e}_{i+\tau}$, are unknown for the future period. 


The decoder transforms a forecasted output pattern into a TS seasonal cycle. The output pattern  predicted by randomized FNN is decoded using the coding variables of the input query pattern, $\mathbf{x}$, using transformed equation \eqref{eq2}:

\begin{equation}\label{eq3}
\widehat{\mathbf{e}} = \widehat{\mathbf{y}}\widetilde{{e}}+\overline{{e}} 
\end{equation}
where $\widehat{\mathbf{e}}$ is the forecasted seasonal sequence, $\widehat{\mathbf{y}}$ is the forecasted output pattern, $\widetilde{e}$ and $\overline{e}$ are the coding variables determined from the TS sequence encoded in query pattern $\mathbf{x}$.

The randomized FNN is composed of $n$ inputs, one hidden layer with $m$ nonlinear nodes, and $n$ outputs. Logistic sigmoid activation functions are employed for hidden nodes. 
The training set is composed of the corresponding input and output patterns: $ \Phi = \left\lbrace (\mathbf{x}_i, \mathbf{y}_i) | \mathbf{x}_i, \mathbf{y}_i \in \mathbb{R}^n,  i = 1, 2,\ldots, N \right\rbrace $. The randomized learning algorithm consists of three steps (we use an improved version of the randomized learning algorithm proposed in \cite{Dud20}):

\begin{enumerate}

\item Generate hidden node parameters. The weights are selected randomly: $a_{t,j} = \sim U(-u, u)$ and biases are calculated from:

\begin{equation}\label{eq4}
b_j = -\mathbf{a}_j^T\mathbf{x}^*_j
\end{equation}
where $j=1, 2 , ..., m$; $t=1, 2, ..., n$; $\mathbf{x}^*_j$ is one of the training x-patterns selected for the $j$-th hidden node at random.  

\item Calculate hidden layer output matrix $\mathbf{H}$.

\item Calculate the output weights: 

	\begin{equation}\label{eq5}
		\boldsymbol{\beta} = \mathbf{H}^+\mathbf{Y}
	\end{equation}
where $ \boldsymbol{\beta} \in \mathbb{R}^{m\times n}$ is a~matrix of output weights, $ \mathbf{Y} \in \mathbb{R}^{N\times n}$ is a~matrix of target output patterns, and $ \mathbf{H}^+ \in \mathbb{R}^{m\times N} $ is the Moore-Penrose generalized inverse of matrix $ \mathbf{H} $.

\end{enumerate}

In the first step, the weights are selected randomly from symmetrical interval $[-u, u]$. The interval bounds, $u$, decide about the steepness of the sigmoids. To make the bounds interpretable, let us express them by the sigmoid slope angle \cite{Dud20}: $u=4 \tan \alpha_{max}$, where $\alpha_{max}$ is the upper bound for slope angles. Hyperparameter $\alpha_{max}$ as well as number of hidden nodes $m$ control the bias-variance tradeoff of the model. Both these hyperparameters should be adjusted to the target function complexity. 

The biases calculated according to \eqref{eq4} ensure the introduction of the steepest fragments of the sigmoid into the input hypercube \cite{Dud20}. These fragments are most useful for modeling target function fluctuations. Thus, there are no saturated sigmoids and wasted nodes in the network.

\section{Ensembling}

An ensemble is composed of $M$ individual learners ($\randnns$). Each ensemble member learns from the training set $\Phi$ using one of the strategies for generating diversity,  $\ensa-\ensf$ described below. The ensemble prediction is an average of individual member predictions $\mathbf{\widehat{e}}_k$:

\begin{equation}\label{eq6}
\mathbf{\widehat{e}}_\text{ens} = \frac{1}{M}\sum_{k=1}^{M} \mathbf{\widehat{e}}_k
\end{equation}

As an ensemble diversity measure, we define the average standard deviation of forecasts produced by the individual learners:

\begin{equation}\label{eq7}
Diversity = \frac{1}{n|\Psi|}\sum_{i \in \Psi}\sum_{t=1}^{n} \sqrt{\frac{1}{M}\sum_{k=1}^{M} (\widehat{E}^k_{i,t} - \overline{\widehat{E}}_{i,t} )^2}
\end{equation}
where $\Psi$ is a test set, $\widehat{E}^k_{i,t}$ is a forecast of the $t$-th element of the $i$-th seasonal sequence produced by the $k$-th learner, and $\overline{\widehat{E}}_{i,t}$ is an average of forecasts produced by $M$ learners.   

An ensemble diversity is generated using one of the following six strategies:
\begin{itemize}
\item [$\ensa$] generates diversity by using different parameters of hidden nodes. For each learner, new weights are randomly selected taking $\alpha_{\max}$ as the upper bound for the sigmoid slope angles. Then, biases are calculated from \eqref{eq4} and output weights form \eqref{eq5}. The diversity level is controlled by $\alpha_{\max}$. For larger $\alpha_{\max}$ we get steeper sigmoids and higher diversity. 

\item [$\ensb$] controls diversity by training individual learners on different subsets of the training set. For
each ensemble member, a random sample from the training set is selected without replacement. The sample size is $N' = \eta N$, where $\eta \in (0 ,1)$ is a diversity parameter. Each learner has the same hidden node parameters. Its output weights are tuned to the training subset. 

\item [$\ensc$] controls diversity by training individual learners on different subsets of features. For each ensemble member the features are randomly sampled without replacement. The sample size is $n' = \kappa n$, where $\kappa \in (0 ,1)$ is a diversity parameter. The ensemble members share the hidden node parameters. Their output weights are tuned to the training set.

\item [$\ensd$] is based on hidden node pruning. The learners are created from the initial $\randnn$ architecture including $m$ hidden nodes. For each learner, $m' = \rho m$ nodes are randomly selected and the remaining are pruned. Output weights $\beta$ are determined anew for each learner. Parameter $\rho \in (0 ,1)$ controls the diversity level.

\item [$\ense$] is based on hidden weight pruning. The learners are created from the initial $\randnn$ architecture including $m$ hidden nodes. For each learner, $p=\lambda mn$ hidden node weights are randomly selected and set to zero. Output weights $\beta$ are determined anew for each learner. Parameter $\lambda \in (0 ,1)$ controls the diversity level.

\item [$\ensf$] generates diversity by noising training data. For each learner the training patterns are perturbed by Gaussian noise as follows: $x_{i,t}=x_{i,t}(1+\zeta_{i,t}), y_{i,t}=y_{i,t}(1+\xi_{i,t})$, where $\zeta_{i,t},\xi_{i,t} \sim N(0, \sigma)$. Standard deviation of the noise, $\sigma$, is a diversity parameter.  Each learner has the same hidden node parameters. Its output weights are tuned to the noised training data. 
\end{itemize}


\section{Experiments and Results}
We evaluate the performance of the proposed ensembles of $\randnns$ on four real-world forecasting problems. These concern forecasting electricity demand for four European countries: Poland (PL), Great Britain (GB), France (FR) and Germany (DE) (data was collected from \url{www.entsoe.eu}).
The hourly electrical load TS express three seasonalities: yearly, weekly and daily (see Fig. 2 in \cite{Dud21}). The data period covers the 4 years from 2012 to 2015. Atypical days such as public holidays were excluded from the data (between 10 and 20 days a year). The forecast horizon $\tau$ is one day, i.e. 24 hours. We forecast the daily load profile for each day of 2015. For each forecasted day, a new training set is created which includes historical pairs of corresponding input and output patterns representing the same days of the week as the pair in the query pattern and forecasted pattern. The number of ensemble members was $M=100$.

\begin{figure}
\centering
\includegraphics[width=1\textwidth]{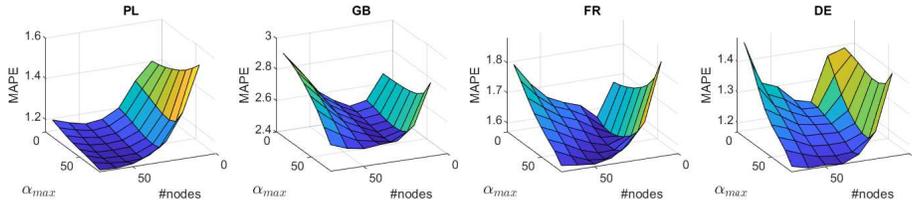}
\caption{MAPE depending on $\randnn$ hyperparameters for $\ensa$.} \label{figE1}
\end{figure}

In the first experiment, to assess the impact of the $\randnns$ hyperparameters on the forecasting accuracy of $\ensa$, we train the $\randnn$ members with $m = 10, 20, ..., 70$ and $\alpha_{\max} = 0^\circ, 10^\circ, ..., 90^\circ$. Fig. \ref{figE1} shows the mean absolute percentage errors (MAPE) depending on the hyperparameters. The optimal values of $m/\alpha_{\max}$ were: $50/ 70^\circ$ for PL, $30/ 60^\circ$ for GB, $40/ 70^\circ$ for FR, and $40/ 80^\circ$ for DE. Based on these results, we select $m=40$ and $\alpha_{\max}= 70^\circ$ for each ensemble variant and dataset, except $\ensa$, where we control diversity by changing $\alpha_{\max}$, and $\ensd$, where we control diversity by changing the number of pruning nodes. In this latter case we use $m=80$ nodes. 

\begin{figure}
\centering
\includegraphics[width=1\textwidth]{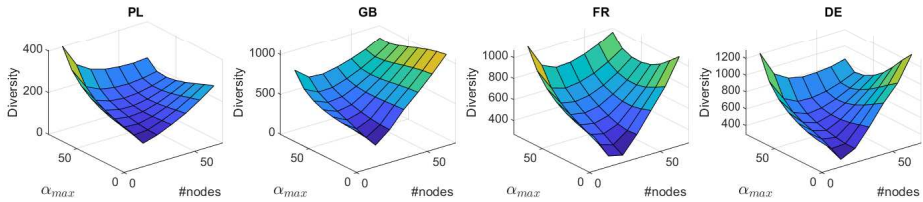}
\caption{Diversity of $\ensa$ depending on $\randnn$ hyperparameters.} \label{figE1a}
\end{figure}

From Fig. \ref{figE1a}, we can assess the impact of the hyperparameters on the diversity \eqref{eq7} of $\ensa$. As expected, greater numbers of hidden nodes and steeper sigmoids (higher $\alpha_{\max}$) cause an increase in diversity.

\begin{figure}
\centering
\includegraphics[width=1\textwidth]{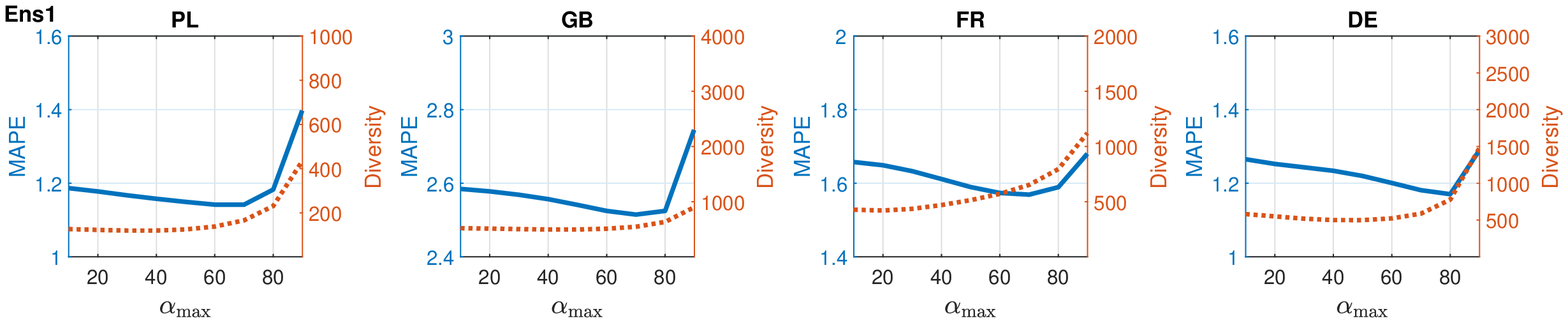}
\includegraphics[width=1\textwidth]{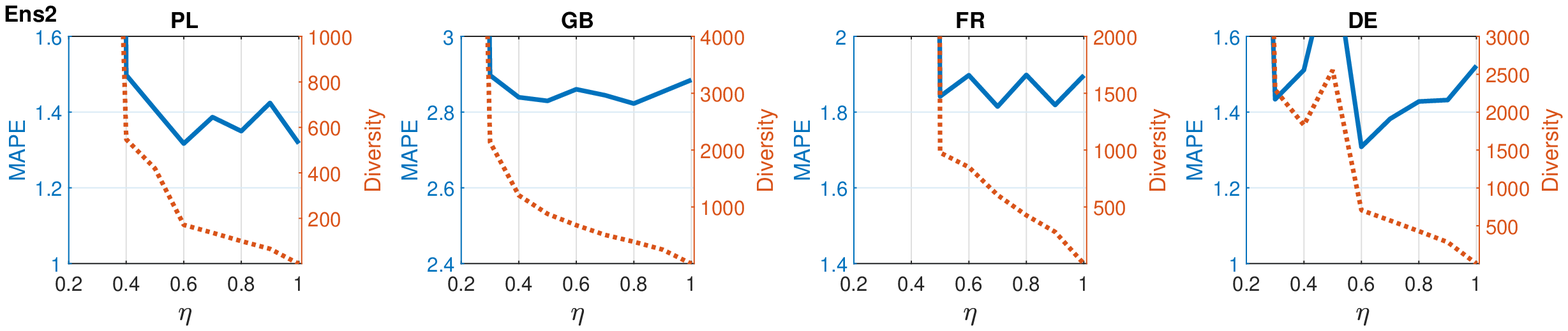}
\includegraphics[width=1\textwidth]{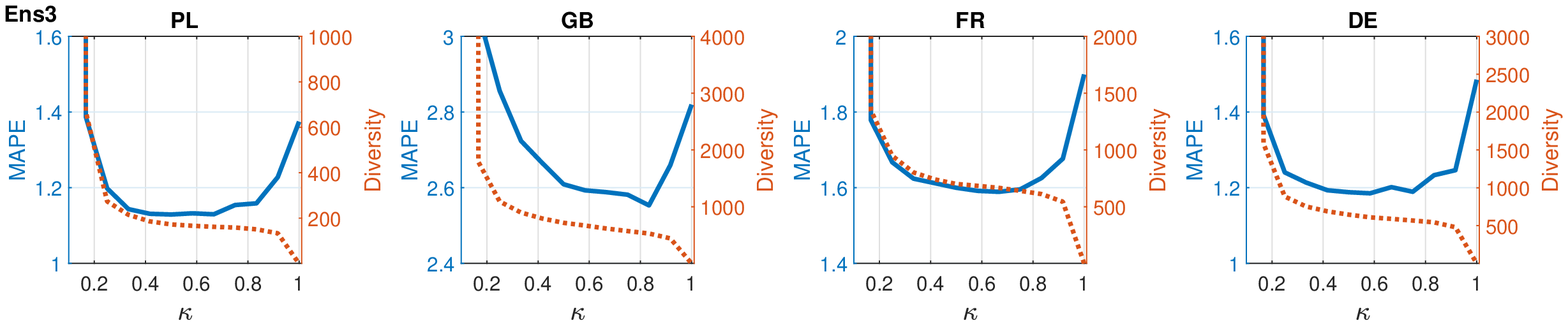}
\includegraphics[width=1\textwidth]{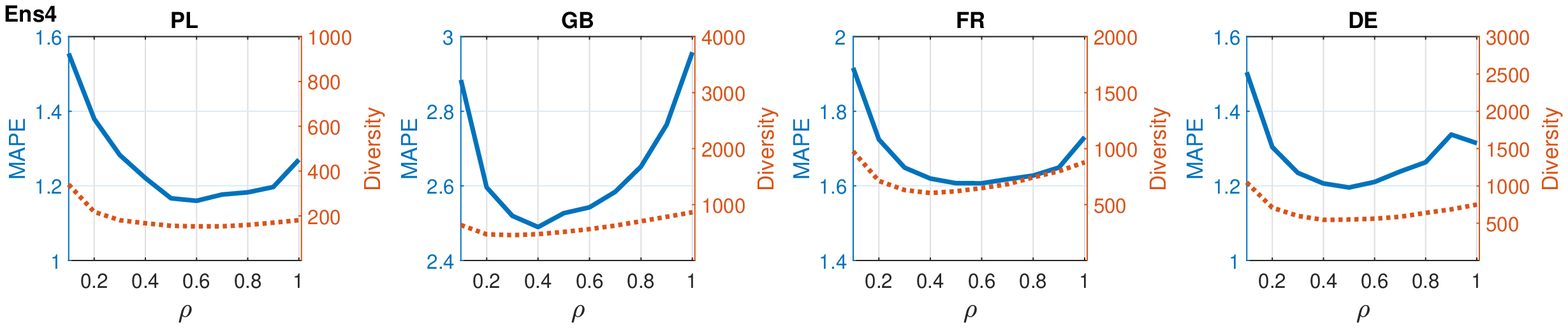}
\includegraphics[width=1\textwidth]{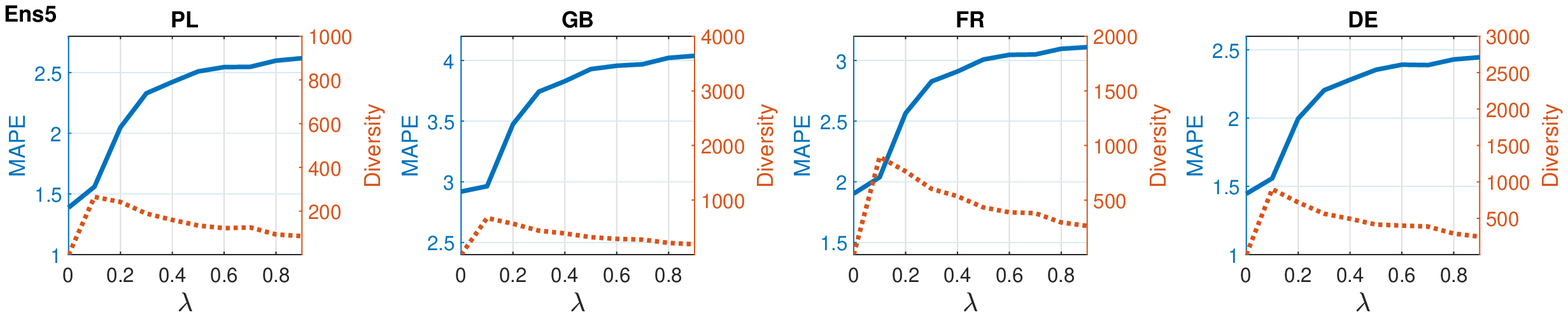}
\includegraphics[width=1\textwidth]{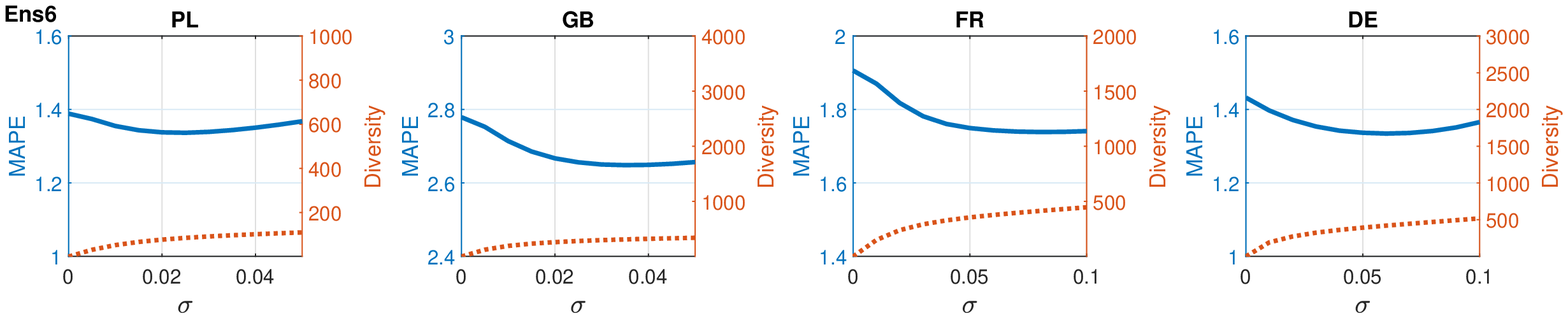}
\caption{MAPE (solid lines) and ensemble diversity (dashed lines) depending on the diversity parameters.} \label{figEns}
\end{figure}

Fig. \ref{figEns} shows MAPE and ensemble diversity depending on the diversity parameters for all ensemble variants and datasets. From this figure, we can conclude that:
\begin{itemize}
\item for $\ensa$ diversity increases with $\alpha_{\max}$. The optimal $\alpha_{\max}$ is around $70^\circ - 80^\circ$ for all datasets. 
\item for $\ensb$ diversity decreases with $\eta$. For lower $\eta$ (small training sets), the diversity as well as MAPE are very high. The MAPE curves have an irregular character, so it is difficult to select the optimal $\eta$ value. The error levels are much higher than for $\ensa$.
\item for $\ensc$ diversity decreases with $\kappa$. MAPE reaches its minima for: $\kappa \in (8/24, 16/24)$ for PL, $\kappa =20/24$ for GB, $\kappa \in (14/24, 18/24)$ for FR, and $\kappa \in (10/24, 18/24)$ for DE. Thus, the best ensemble solutions use around $40^\circ - 84^\circ$ of features.   
\item for $\ensd$ diversity changes slightly with $\rho$. MAPE reaches its minima for: $\rho \in (40/80, 48/80)$ for PL, $\rho =32/80$ for GB, $\rho \in (40/80, 48/80)$ for FR, and $\rho =40/80)$ for DE. Thus, the best choice for the number of selected nodes is around $m'=40$.
\item for $\ense$ diversity has its maximum for $\lambda=0.1$ (10\% of hidden weights are set to 0). MAPE increases with $\lambda$, having its lowest values for $\lambda=0$ (no weight pruning). Thus, any weight pruning makes the error higher. Increasing the hidden node number to 80 did not improve the results.
\item for $\ensf$ we observe low sensitivity of diversity as well as MAPE to diversity parameter $\sigma$. For smaller $\sigma$, the error only slightly decreases as $\sigma$ increases. Then, for higher $\sigma$, the error starts to increase.  

\end{itemize}

\begin{figure}
\centering
\includegraphics[width=0.24\textwidth]{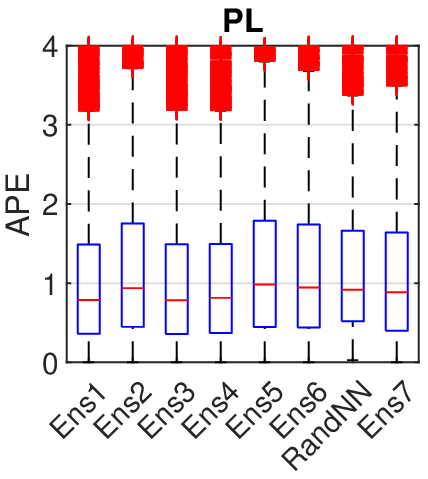}
\includegraphics[width=0.24\textwidth]{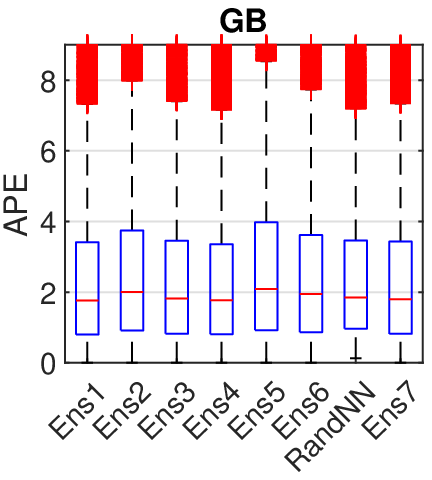}
\includegraphics[width=0.24\textwidth]{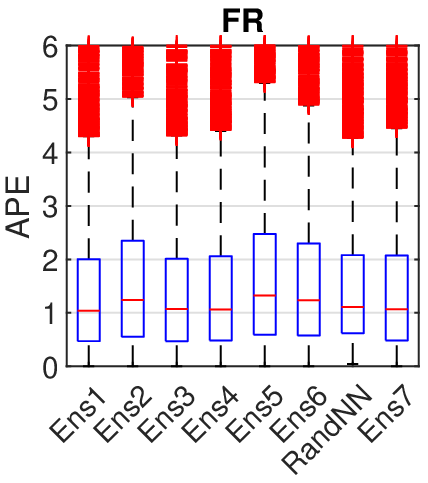}
\includegraphics[width=0.24\textwidth]{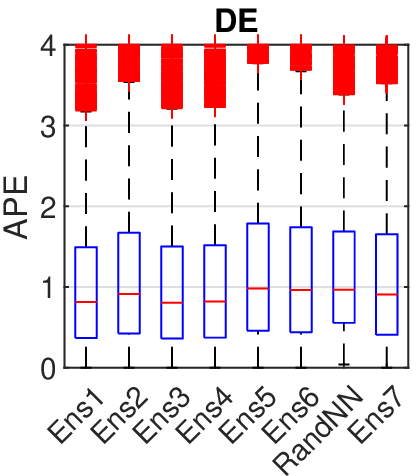}
\caption{Distribution of APE.} \label{fig5}
\end{figure}

Table \ref{tab1} shows the quality metrics of the forecasts for the optimal values of the diversity parameters: MAPE, median of APE, root mean square error (RMSE), mean percentage error (MPE), and standard deviation of percentage error (Std(PE)) as a measure of the forecast dispersion. For comparison, the results for a single $\randnn$ are shown. In this case, for each forecasting task (each day of 2015) the hyperparameters of $\randnn$, $m$ and $u$, were optimized using grid search and 5-fold cross-validation \cite{Dud21}. The results for single $\randnn$ shown in Table \ref{tab1} are averaged over 100 independent training sessions. $\ensg$ shown in Table \ref{tab1} is an ensemble of these 100 runs (calculated from \eqref{eq6}). $\ensg$ is similar to $\ensa$. The only difference is that for $\ensg$ the $\randnn$ hyperparameters were optimized for each forecasting task, and for $\ensa$ we set $m=40$ and $\alpha_{\max} = 70^\circ$ for all forecasting tasks. Distributions of APE are shown in Fig. \ref{fig5}. 

\begin{table}[htbp]
  \begin{center}
\setlength{\tabcolsep}{4pt}
  \caption{Forecasting results.} \label{tab1}
    \begin{tabular}{llcccccccc} 
    \toprule
      & & \multicolumn{1}{r}{$\ensa$} & \multicolumn{1}{r}{$\ensb$} & \multicolumn{1}{r}{$\ensc$} & \multicolumn{1}{r}{$\ensd$} & \multicolumn{1}{r}{$\ense$} & \multicolumn{1}{r}{$\ensf$} & \multicolumn{1}{r}{$\randnn$} & \multicolumn{1}{r}{$\ensg$} \\
    \midrule
    \textbf{PL} & MAPE & 1.14  & 1.32  & 1.13  & 1.16  & 1.39  & 1.34  & 1.32  & 1.24 \\
    & Median(APE) & 0.79  & 0.94  & 0.78  & 0.81  & 0.98  & 0.94  & 0.93  & 0.89 \\
    & RMSE  & 304 & 342 & 299 & 313 & 380 & 351 & 358 & 333 \\
    & MPE   & 0.31  & 0.43  & 0.31  & 0.25  & 0.36  & 0.45  & 0.40  & 0.40 \\
    & Std(PE) & 1.67  & 1.88  & 1.64  & 1.72  & 2.07  & 1.94  & 1.94  & 1.80 \\
    \midrule
    \textbf{GB} & MAPE & 2.51  & 2.74  & 2.55  & 2.49  & 2.92  & 2.65  & 2.61  & 2.52 \\
    &Median(APE) & 1.76  & 2.01  & 1.82  & 1.77  & 2.09  & 1.95  & 1.88  & 1.80 \\
    &RMSE  & 1151 & 1205 & 1160 & 1131 & 1325 & 1167 & 1187 & 1147 \\
    &MPE   & -0.53 & -0.83 & -0.53 & -0.53 & -0.55 & -0.65 & -0.61 & -0.61 \\
    &Std(PE) & 3.48  & 3.64  & 3.51  & 3.42  & 4.02  & 3.56  & 3.57  & 3.44 \\
   \midrule
    \textbf{FR} & MAPE & 1.57  & 1.80  & 1.59  & 1.61  & 1.90  & 1.74  & 1.67  & 1.60 \\
    &Median(APE) & 1.04  & 1.24  & 1.07  & 1.06  & 1.33  & 1.24  & 1.15  & 1.07 \\
    &RMSE  & 1378 & 1505 & 1398 & 1402 & 1597 & 1459 & 1422 & 1385 \\
    &MPE   & -0.28 & -0.26 & -0.25 & -0.31 & -0.33 & -0.27 & -0.42 & -0.42 \\
    &Std(PE) & 2.53  & 2.80  & 2.58  & 2.57  & 2.92  & 2.72  & 2.60  & 2.53 \\
    \midrule
    \textbf{DE} & MAPE & 1.18  & 1.31  & 1.19  & 1.20  & 1.45  & 1.33  & 1.38  & 1.29 \\
    &Median(APE) & 0.81  & 0.91  & 0.80  & 0.82  & 0.98  & 0.96  & 0.96  & 0.91 \\
    &RMSE  & 1077 & 1155 & 1111 & 1097 & 1423 & 1159 & 1281 & 1206 \\
    &MPE   & 0.11  & 0.03  & 0.10  & 0.09  & 0.02  & 0.17  & 0.14  & 0.14 \\
    &Std(PE) & 1.89  & 2.01  & 1.96  & 1.92  & 2.44  & 2.04  & 2.22  & 2.11 \\
    \bottomrule
    \end{tabular}%
  \label{tab1}%
  \end{center}
\end{table}%

\begin{figure}
\centering
\includegraphics[width=0.227\textwidth]{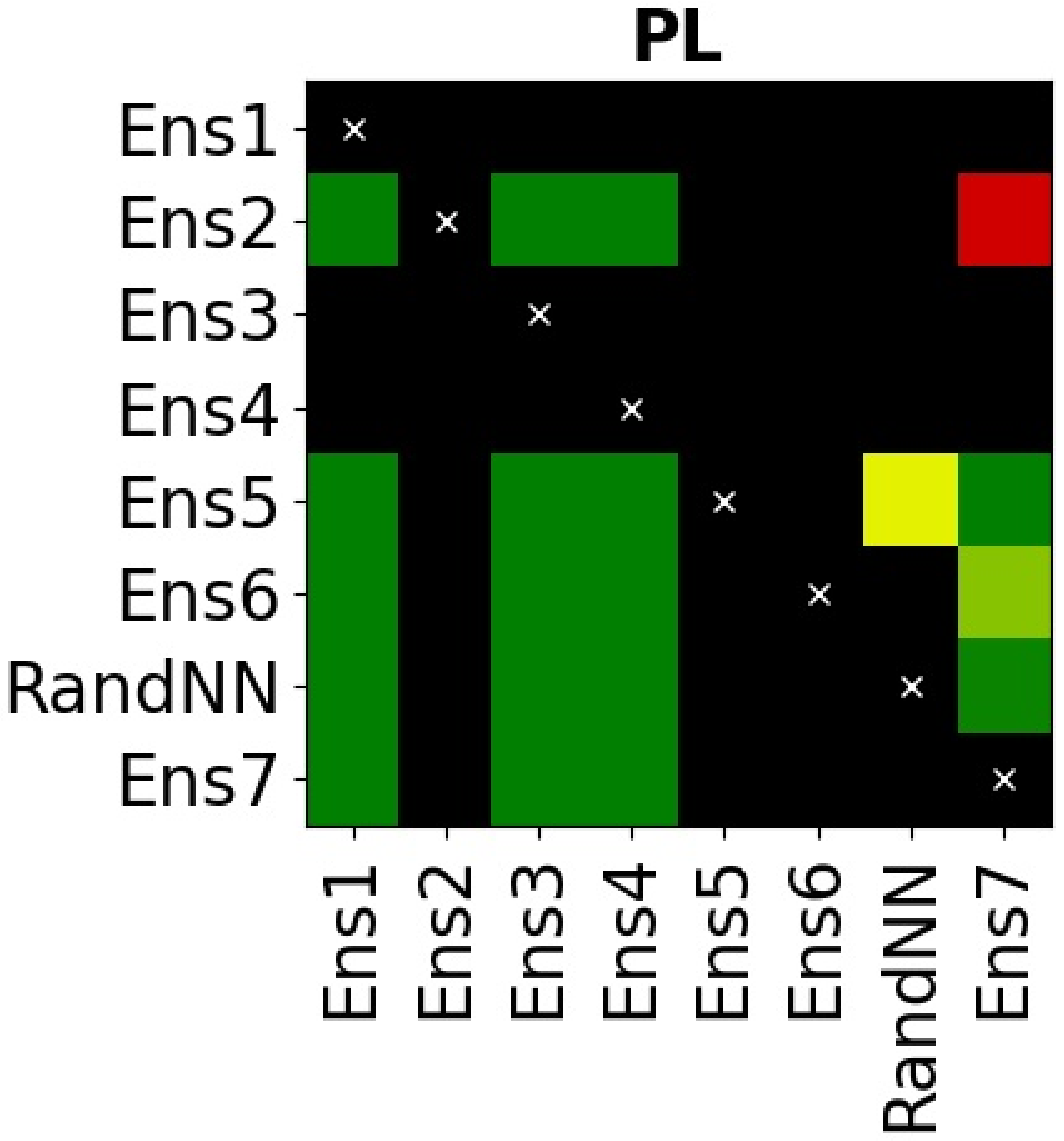}
\includegraphics[width=0.227\textwidth]{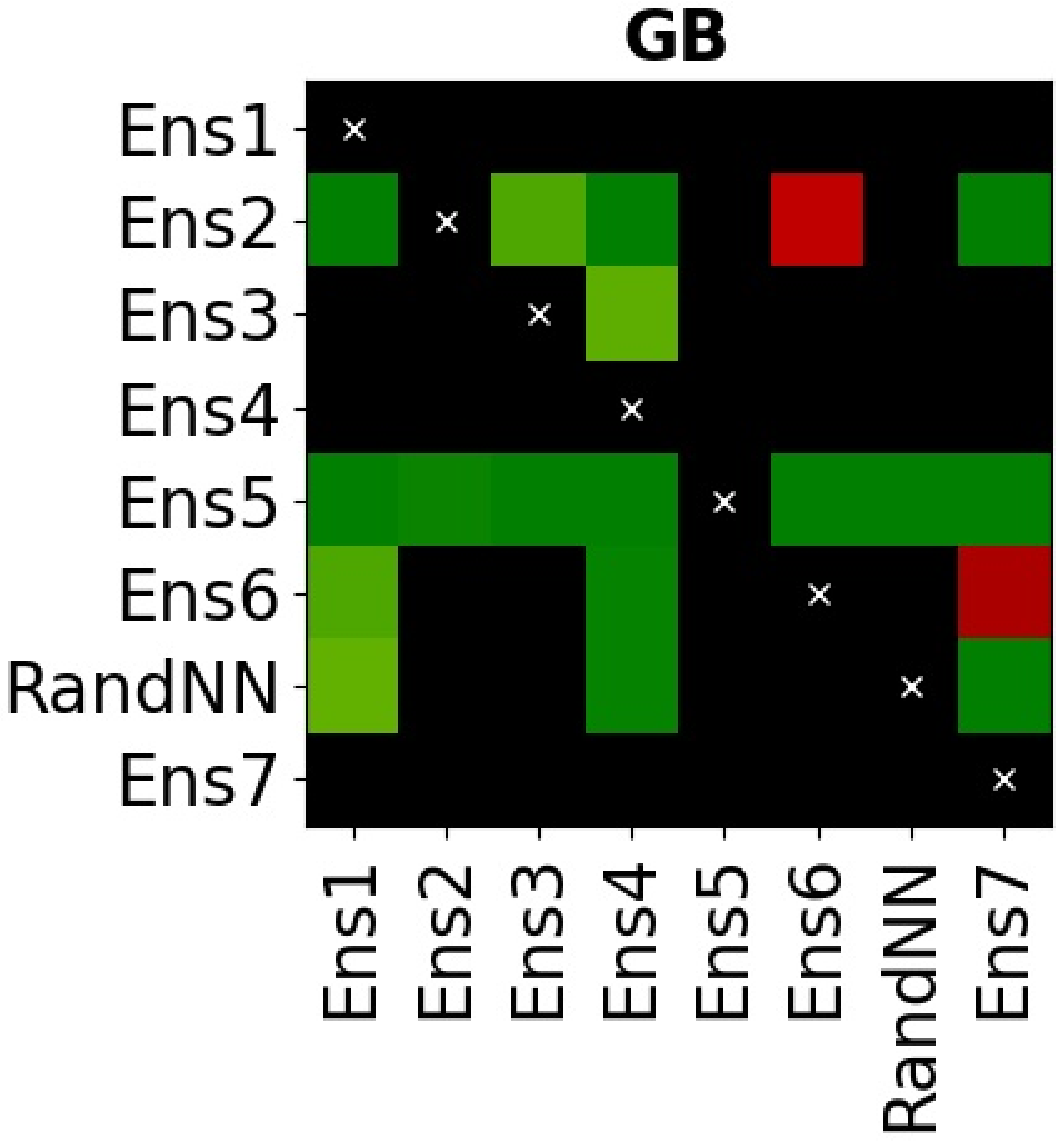}
\includegraphics[width=0.227\textwidth]{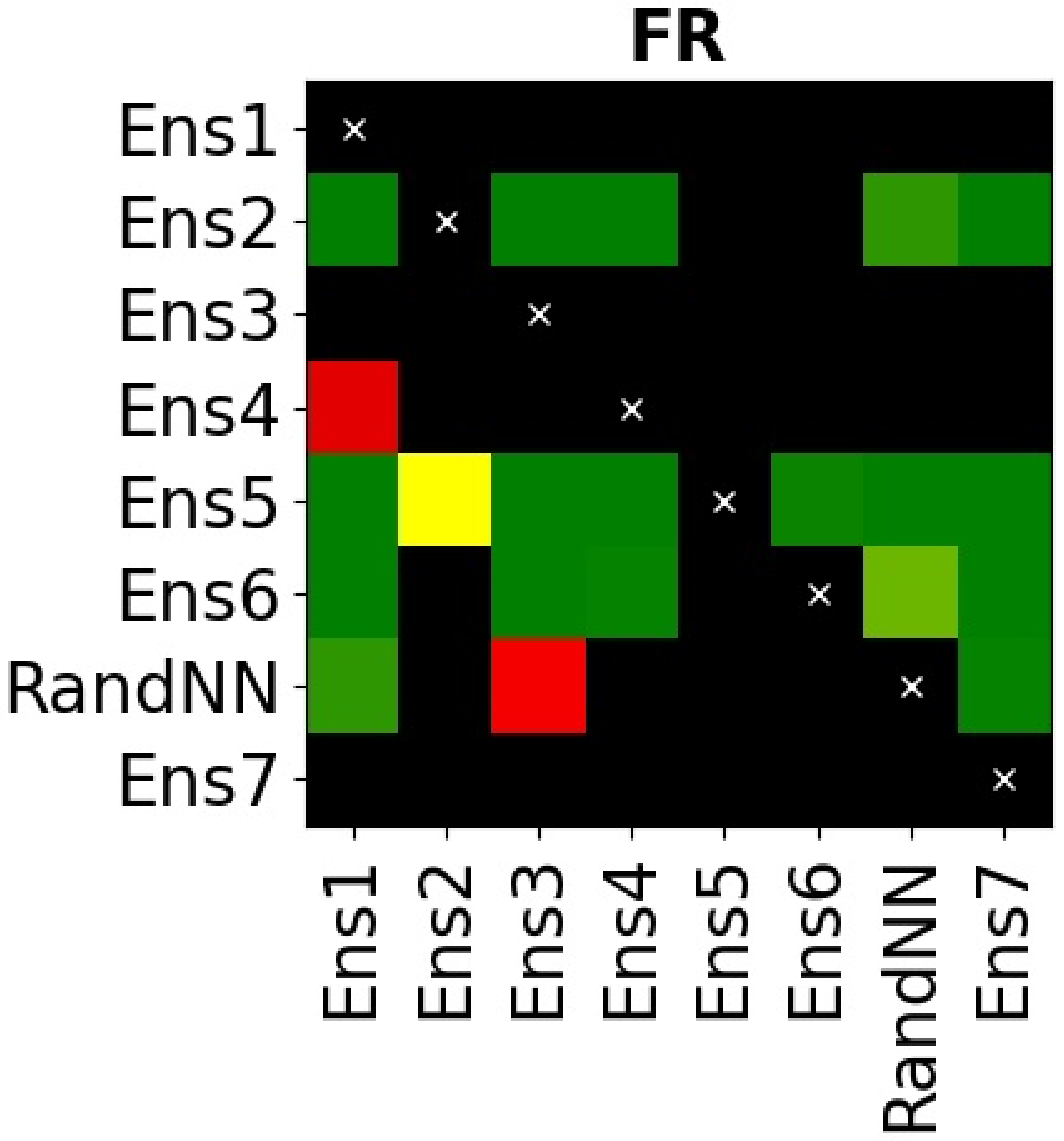}
\includegraphics[width=0.29\textwidth]{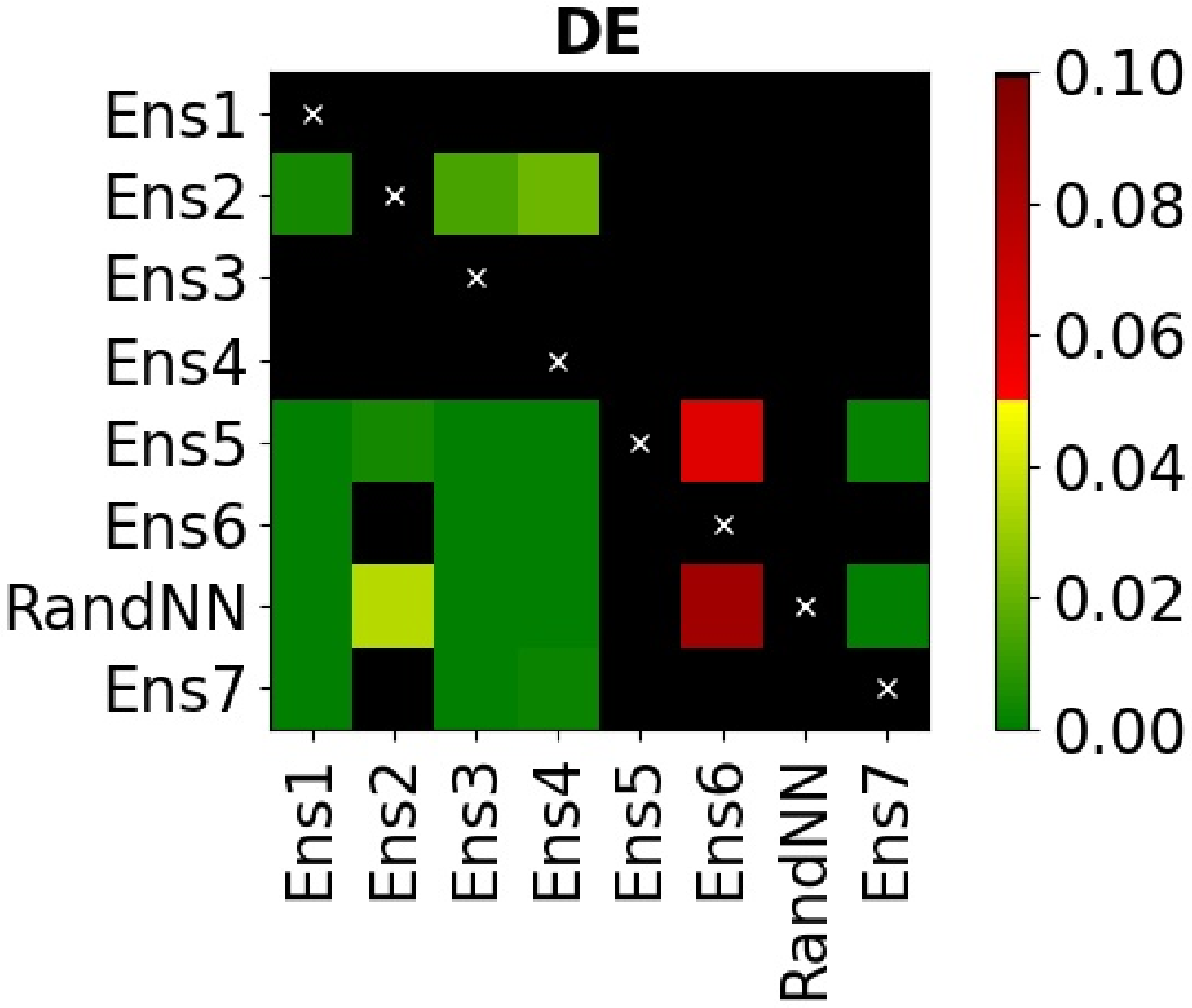}
\caption{Results of the Giacomini-White test for the proposed $\randnn$ ensembles.} \label{fig6}
\end{figure}

To confirm that the differences in accuracy between ensemble variants are statistically significant, we performed a one-sided Giacomini-White test for conditional predictive ability \cite{Gia06}.  
This is a pairwise test that compares the forecasts produced by different models. We used a Python implementation of the Giacomini-White test in multivariate variant from \cite{epf,Lag21}.
Fig. \ref{fig6} shows results of the Giacomini-White test. The resulting plots are heat maps representing the obtained $p$-values. The closer they are to zero
the significantly more accurate the forecasts produced by the model on the $X$-axis are than the forecasts produced by the model on the $Y$-axis.
The black color indicates $p$-values larger than 0.10. 

From the results presented in Table \ref{tab1} and Figs. \ref{fig5} and \ref{fig6} we can conclude that the most accurate ensembles are $\ensa$, $\ensc$, and $\ensd$. For PL and DE they significantly outperformed all other models including single $\randnn$ and $\ensg$, although, for GB and FR, $\ensg$ can compete with them in terms of accuracy. Note that the APE distributions for $\ensa$, $\ensc$, and $\ensd$ shown in a simplified form in Fig. \ref{fig5} are very similar. The worst ensemble solution was $\ense$ based on weight pruning. It was beaten by all other models for all datasets. $\ensb$ and $\ensf$ were only slightly better in terms of accuracy than $\ense$.

MPE shown in Table \ref{tab1} allows us to assess the bias of the forecasts. A positive value of MPE indicates underprediction, while a negative value indicates overprediction. As can be seen from Table \ref{tab1}, for PL and DE all the models underpredicted, whilst for GB and FR they overpredicted. The forecasts produced by the three best ensemble models were less biased than the forecast produced by a single $\randnn$ and $\ensg$.  

In the next experiment, we compare $\ensa$ performance (one of the best ensemble solutions) with that of other models based on classical statistical methods and ML methods. The baseline models that we use in our comparative studies are outlined below (see \cite{Dud21a} for details). Their hyperparameters were selected on the training set in grid search procedures.

\begin{itemize}
    \item \textsc{naive} -- naive model: $\widehat{\mathbf{e}}_{i+\tau}=\mathbf{e}_{i+\tau-7}$,
    \item \textsc{arima} -- autoregressive integrated moving average model,
    
    \item \textsc{ets} -- exponential smoothing model,
	\item \textsc{prophet} -- a modular additive regression model with nonlinear trend and seasonal components \cite{Tay18},
	\item \textsc{mlp} -- perceptron with a single hidden layer and sigmoid nonlinearities,
	\item \textsc{svm} -- linear epsilon insensitive support vector machine ($\epsilon$-SVM) \cite{Pel21},
	\item \textsc{anfis} -- adaptive neuro-fuzzy inference system,
	\item \textsc{lstm} -- long short-term memory,
	\item \textsc{fnm} -- fuzzy neighborhood model,
	\item \textsc{n-we} -- Nadaraya–Watson estimator,
	\item \textsc{grnn} -- general regression NN.
\end{itemize}

\begin{figure}
\centering
\includegraphics[width=0.227\textwidth]{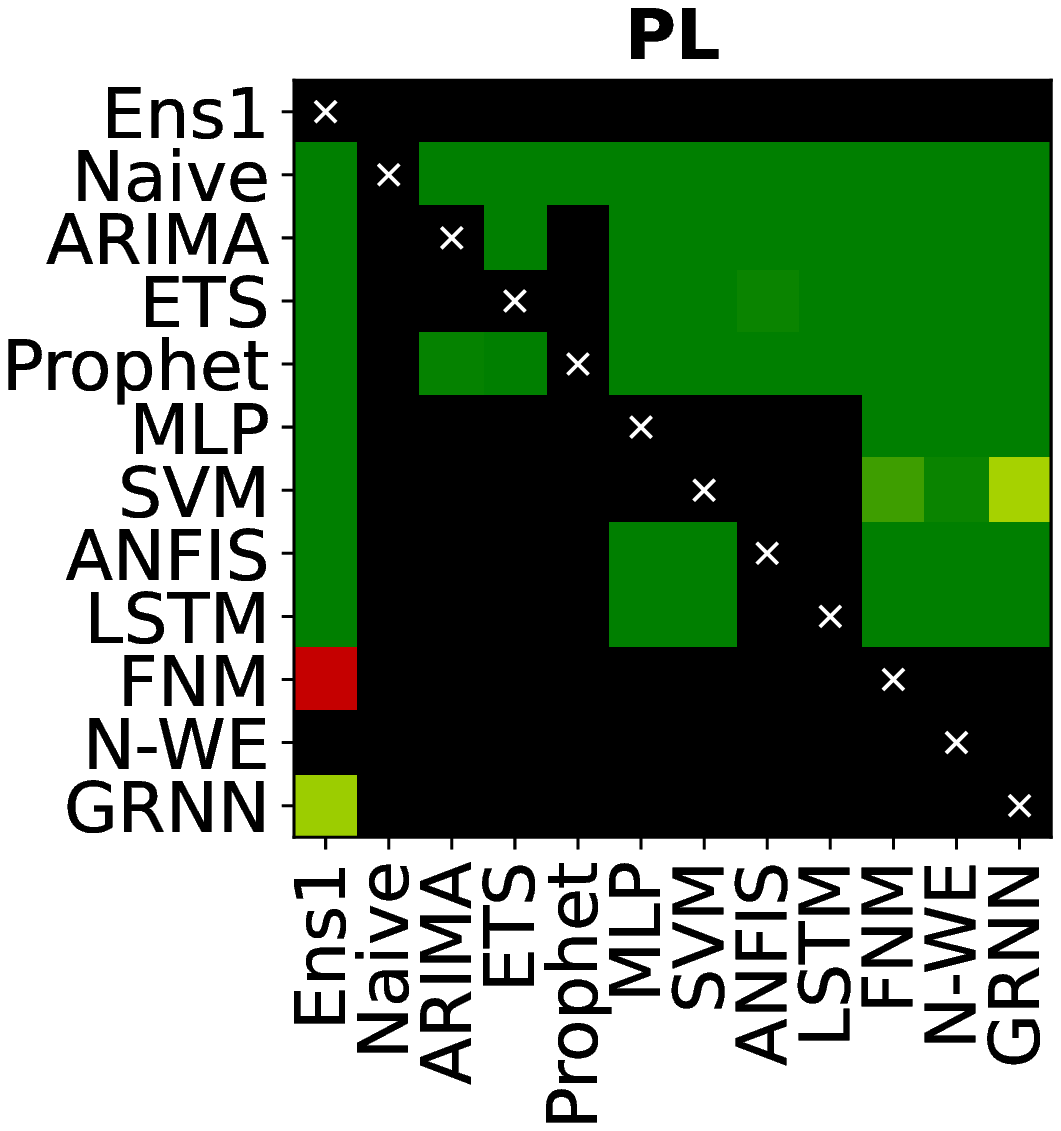}
\includegraphics[width=0.227\textwidth]{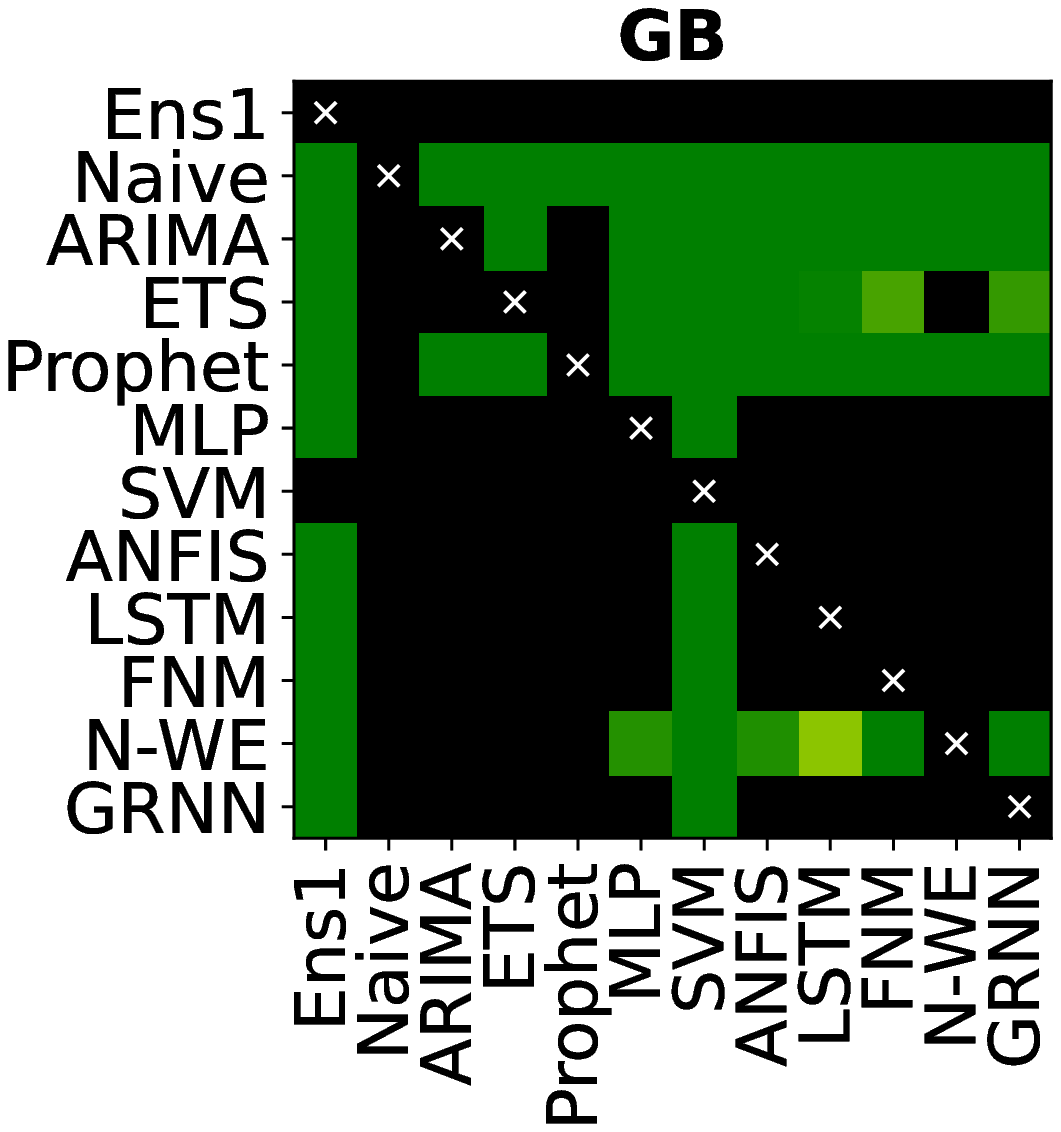}
\includegraphics[width=0.227\textwidth]{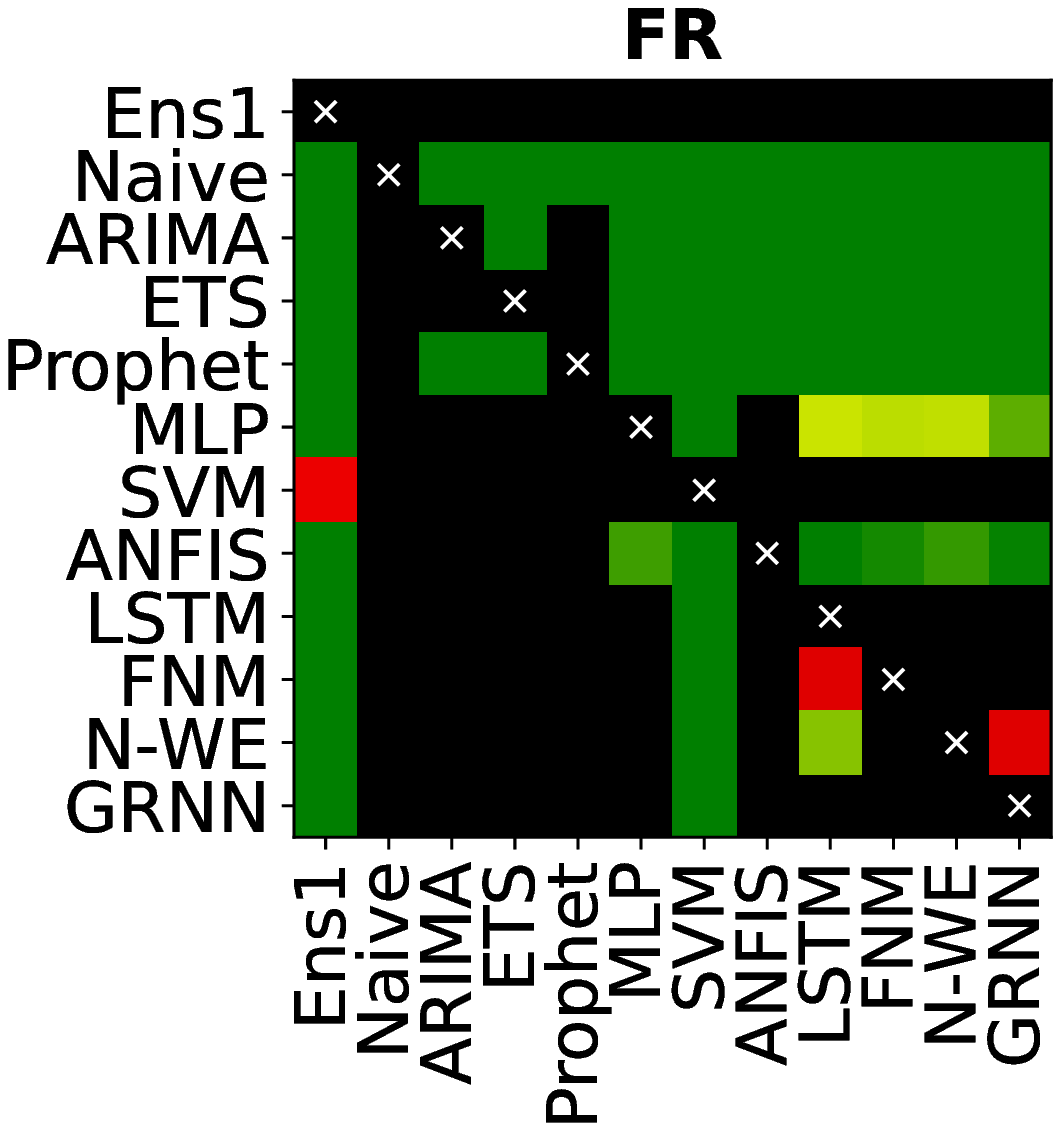}
\includegraphics[width=0.29\textwidth]{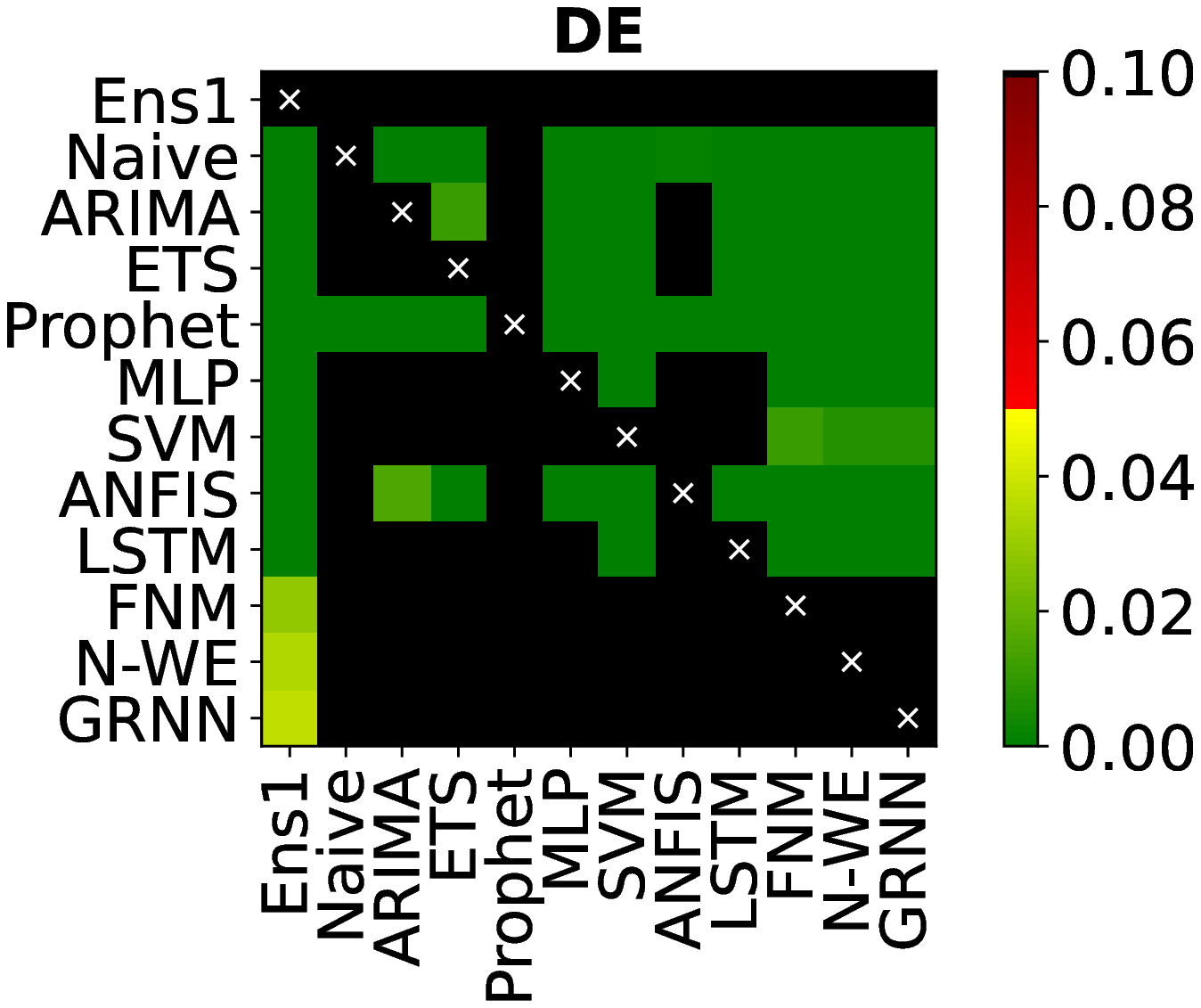}
\caption{Results of the Giacomini-White test for the proposed $\ensa$ and baseline models.} \label{fig7}
\end{figure}

Table \ref{tab2} shows MAPE for $\ensa$ and the baseline models. 
Note that for each dataset, $\ensa$ returned the lowest MAPE. To confirm the statistical significance of these findings, we performed a Giacomini-White test. The test results depicted in Fig. \ref{fig7}, clearly show that $\ensa$ outperforms all the other models in terms of accuracy. 
Only in two cases out of 44 were the baseline models close to $\ensa$ in terms of accuracy, i.e. N-WE for PL data and SVM for GB data.

\begin{table}[htbp]
\begin{center}
\setlength{\tabcolsep}{2.6pt}
   \caption{MAPE for $\ensa$ and baseline models.}
    \begin{tabular}{lccccccccccccc}
    \toprule
          & $\ensa$  & \textsc{naive} & \textsc{arima} & \textsc{ets}   & \textsc{prophet} & \textsc{mlp}   & \textsc{svm}   & \textsc{anfis} & \textsc{lstm}  & \textsc{fnm}   & \textsc{n-we}  & \textsc{grnn} \\
    \midrule
    \textbf{PL} & 1.14  & 2.96  & 2.31  & 2.14  & 2.63  & 1.39  & 1.32  & 1.64  & 1.57  & 1.21  & 1.19  & 1.22 \\
    \textbf{GB}  & 2.51  & 4.80  & 3.50  & 3.19  & 4.00  & 2.84  & 2.54  & 2.80  & 2.92  & 3.02  & 3.12  & 3.01 \\
    \textbf{FR}  & 1.57  & 5.53  & 3.00  & 2.79  & 4.71  & 1.93  & 1.63  & 2.12  & 1.81  & 1.84  & 1.86  & 1.81 \\
    \textbf{DE}  & 1.18  & 3.13  & 2.31  & 2.10  & 3.23  & 1.58  & 1.38  & 2.48  & 1.57  & 1.30  & 1.29  & 1.30 \\
    \bottomrule
    \end{tabular}%
  \label{tab2}%
  \end{center}
\end{table}%

\section{Conclusion}

Challenging forecasting problems, such as forecasting nonstationary TS with multiple seasonality, need sophisticated models. In this work, to deal with multiple seasonal periods, we employ a pattern representation of the TS, in order to simplify the relationship between input and output data and make the problem easier to solve using simple regression models such as randomized NNs. $\randnns$ have three distinct advantages: a simple single-hidden layer architecture, extremely fast learning (no gradient-based learning, no problems with vanishing and exploding gradients), and ease of implementation (no complex optimization algorithms, no additional mechanisms such as dilation, attention, and residual connections). $\randnn$ produces a vector output, which means the model is able to forecast the time series sequence at once. 

We propose an ensemble of $\randnns$ with six different methods of managing diversity. Among these, the most promising ones turned out to be: learning the ensemble members using different parameters of hidden nodes ($\ensa$), training individual learners on different subsets of features ($\ensc$), and hidden node pruning ($\ensd$). These three methods brought comparable results in an experimental study involving four real-world forecasting problems (short-term load forecasting). $\ensa$ has the additional advantage of having only two hyperparameters to tune, i.e. the number of hidden nodes and the interval bounds for random weights. In the comparative study, $\ensa$ significantly outperformed both classical statistical methods and machine learning methods.

In our further work, we plan to introduce an attention mechanism into our randomization-based forecasting models to select training data and develop probabilistic forecasting models based on $\randnns$.

%
%
%

\end{document}